\documentclass{article}

\usepackage{microtype}
\usepackage{graphicx}
\usepackage{subfigure}
\usepackage{subcaption}
\usepackage{booktabs} % for professional tables

\usepackage{hyperref}

\usepackage{algpseudocode}
\usepackage{algorithm}

\usepackage[accepted]{icml2025}

\usepackage{amsmath}
\usepackage{amssymb}
\usepackage{mathtools}
\usepackage{amsthm}

\usepackage[capitalize,noabbrev]{cleveref}

\theoremstyle{plain}

\theoremstyle{definition}

\theoremstyle{remark}

\usepackage[textsize=tiny]{todonotes}

\icmltitlerunning{Tracking World States with Language Models: State-Based Evaluation Using
Chess}

\begin{document}

\twocolumn[
\icmltitle{Tracking World States with Language Models: State-Based Evaluation Using Chess}



\icmlsetsymbol{equal}{*}

\begin{icmlauthorlist}
\icmlauthor{Romain Harang}{utokyo}
\icmlauthor{Jason Naradowsky}{utokyo}
\icmlauthor{Yaswitha Gujju}{utokyo}
\icmlauthor{Yusuke Miyao}{utokyo}
\end{icmlauthorlist}

\icmlaffiliation{utokyo}{Department of Computer Science, University of Tokyo, Bunkyo, Japan}
%\icmlaffiliation{square}{Square Enix: AI & Arts Alchemy, Japan}

\icmlcorrespondingauthor{Romain Harang}{romain-harang@g.ecc.u-tokyo.ac.jp}

\icmlkeywords{State Tracking, Evaluation, Metrics}

\vskip 0.3in
]



\printAffiliationsAndNotice{}  % leave blank if no need to mention equal contribution

\begin{abstract}
Large Language Models (LLMs) exhibit emergent capabilities in structured domains, suggesting they may implicitly internalize high-fidelity representations of world models. While probing techniques have shown promising signs of this in scientific and game-based settings, they rely on model-specific internal activations, which limit interpretability and generalizability. In this work, we propose a model-agnostic, state-based evaluation framework using chess as a benchmark to assess whether LLMs preserve the semantics of structured environments. Our method analyzes the downstream legal move distributions (state affordances) to estimate semantic fidelity between predicted and actual game states. This approach offers a more meaningful evaluation than conventional string-based metrics by aligning more closely with the strategic and rule-governed nature of chess. Experimental results demonstrate that our metrics capture deficiencies in state-tracking, highlighting limitations of LLMs in maintaining coherent internal models over long sequences. Our framework provides a robust tool for evaluating structured reasoning in LLMs without requiring internal model access, and generalizes to a wide class of symbolic environments.
\end{abstract}

\section{Introduction}

Large Language Models (LLMs), consisting of billions of parameters trained on massive text corpora, have demonstrated capabilities far beyond their original next-token prediction task. Recent studies suggest that this enhanced ability arises from their implicit recovery of high-fidelity representations of structured domains embedded within their training data. This implicit modeling aligns closely with the concept of a “world model,” defined in \citep{worldmodelHa} for Neural Networks in general. In our case, it can be understood as a representation where an environment can be summarized by a finite set of states and rules governing transitions between them—effectively modeled as a deterministic finite automaton (DFA) \citep{vafa2024evaluating}.

Language models have shown promise in recovering such world models purely from sequential data in complex scientific domains like protein design, genetics, and chemistry \citep{chowdhury2022single, lin2023evolutionary, benegas2023dna, jablonka2024leveraging}. This ability offers a powerful alternative to explicitly constructing detailed representations of complex environments, highlighting the capacity of language models to extract rich domain knowledge solely from sequences. However, these successes rest on a critical assumption: that the language model has genuinely internalized the underlying world model. This raises a fundamental question—how can we reliably determine whether a sequence model has truly learned the domain’s structure?

A common strategy for evaluating whether a model has internalized a world model involves probing its internal neural representations to see if they can reconstruct real-world states \cite{hewitt2019designing, li2021implicit, abdou2021can, jin2024emergent, li2023emergent}. For example, \citep{toshniwal2022chess} and \cite{li2023emergent} evaluate whether sequence models trained on board game transcripts, such as chess and Othello, have internalized the underlying game rules. However, these probing-based methods rely heavily on accessing and interpreting internal model states, which can be model-specific, opaque, and challenging to generalize. Furthermore, evaluating the quality of generated move sequences in chess remains difficult because most existing metrics focus on syntactic-level comparisons, such as exact match, edit distance, Levenshtein distance, or direct board state comparisons. While straightforward to compute, these metrics fail to capture the strategic and semantic richness of chess, where moves vary greatly in their impact—some drastically influence the course of the game, while others are strategically neutral or suboptimal.

Motivated by these limitations, our work proposes a model-agnostic, sequence-based evaluation framework that directly examines the model’s generated outputs to determine whether they preserve the semantic properties of the original game state. We accomplish this by analyzing the space of valid action sequences that can unfold from a given position and evaluating the similarity between predicted and actual states based on the moves they enable. Although computationally more demanding than string-based metrics, our approach provides a richer and more informative signal, capturing whether an LLM’s output retains the tactical and strategic affordances inherent in real chess positions. This framework enables robust inference and also applies broadly across different model architectures without requiring probing of internal representations.

\section{Background}

We base our framework on Finite State Automata (FSA), which offer a natural way to model state tracking in structured environments. By comparing the sets of valid continuations from different action sequences, we capture a semantics-aware notion of similarity.

\subsection{State tracking}

% We start from a state $S\in\mathcal{S}$ that is defined based on the problem at hand, this state can be modified by actions $a\in\Sigma$. Certain actions will be permitted depending on $S$ so we define the subset of actions permitted under $S$ $\Sigma_S\subset\Sigma$. When the starting state $S_0$ is defined and using the transition function $\delta:\mathcal{S}\times\Sigma\rightarrow\mathcal{S}$ we can reach a state $S_n$ from $S_0$ using $n$ actions $a_i\in\Sigma_{S_{i-1}}$. Then $S_j=\delta(S_{j-1},a_j)$ is the recurrent formula that allows us to from $S_0$ to $S_n$, we can also represent $S_n$ as the action sequence $s=(a_1,..., a_n)$.\\
% From this we understand that an action sequence (assuming a starting state) has an unique associated state. And therefore, this state can be taken from it as long as we have access to the transition function $\delta$ or how actions modify states.\\
% A way to test if a model develops a world model is to check its state tracking abilities. Starting from a task akin to language modeling we define over an alphabet $\Sigma$ elements $s\in\Sigma$ then a sequence will be defined as $s_1s_2...s_n$ that we will also note $\Pi_{i=1}^ns_i$. Elements in a sequence do not commute arbitrarily and elements can be added to a sequence if they belong to the accepted subset of $\Sigma$ by the sequence. In the cases we are 
\paragraph{Finite State Automata}
We define a Finite State Automaton (FSA) as the tuple $
\mathcal{A} = (\mathcal{S}^*, \Sigma, \delta, S_0) $
We define the setting as follows. Let $\mathcal{S}$ denote the finite set of valid states. We augment this set to $\mathcal{S}^* = \mathcal{S} \cup \{0\}$, where $0$ represents a special error or sink state that captures invalid transitions. The input alphabet, denoted by $\Sigma$, is the finite set of actions that can be applied to states. The process begins from an initial state $S_0$, which belongs to $\mathcal{S}$. Transitions between states are governed by a function $\delta$, which maps a pair consisting of a current state (from $\mathcal{S}^*$) and an action (from $\Sigma$) to a new state in $\mathcal{S}^*$.
Each state $S \in \mathcal{S}$ has an associated set of permitted actions, denoted $\Sigma_S \subseteq \Sigma$. 
\paragraph{Transition Sequences.}
Given an action sequence $s = (a_1, \dots, a_n) \in \Sigma^{\mathbb{N}}$ and the initial state $S_0$, the resulting sequence of states $(S_1, \dots, S_n)$ is defined recursively by
$
S_i = \delta(S_{i-1}, a_i), \quad \text{for } i = 1, \dots, n.
$
%\color{blue}
We can also extend the transition function's definition over a sequence of moves rather than a single move, for example here: $S_n=\delta(S_0,s)$\color{black}. Thus, an action sequence $s$ uniquely determines the final state $S_n$, provided the transition function $\delta$ and initial state $S_0$ are known. In this sense, the final state is a function of the sequence.

\paragraph{Evaluating State Tracking.}
To assess whether a model has internalized a form of world-modeling or state tracking, we examine its performance on structured action sequences. The model only uses action sequences as inputs, without access to the explicit underlying state structure. If generating valid continuations requires an implicit estimate of the current state, we may conclude that the model has tracked the relevant state information. This framework allows us to test a model’s state awareness: if it can accurately predict or generate valid next actions under the constraints imposed by $\delta$, then it implicitly represents a state in a manner consistent with a world model.

%\color{blue}
\subsection{Existing approaches}

Existing approaches are typically developed with the specific task in mind. In the case of chess, which we consider in our experiments, the two main evaluation metrics are board accuracy and edit distance \citep{feng2023chessgptbridgingpolicylearning}. However, these techniques do not take into account the affordances associated with a particular board state. As a result, they can yield "false positives" in certain scenarios. For example, if two states differ only by the removal of a king, the edit distance might be as low as $1$ and the board accuracy could exceed $98$\%, even though the resulting state is nonsensical from an affordance perspective.

While we expect the metrics derived from our approach to be correlated with these baseline metrics, we argue that they more directly capture the LLM's understanding of the task. Specifically, a higher score in our framework consistently corresponds to a state that is semantically closer to the ground truth, unlike traditional metrics. Although it is possible to introduce heuristic weighting into edit distance or accuracy, a key advantage of our method is that it requires no prior knowledge of the task. It can be applied directly in any setting, as long as $\delta$ and $\Sigma$ are defined.

\color{black}

\section{Main idea}

\subsection{State reconstruction task}
We define a complementary evaluation objective, the state reconstruction task, which assesses whether a model can explicitly generate the current environment state after observing a sequence of actions. In domains where states admit a textual representation (e.g., algebraic notation in chess), we prompt the model to produce this representation given only the preceding action sequence. While failure to reconstruct the state does not conclusively imply that state tracking has not occurred (e.g., due to formatting or generation noise), successful reconstruction provides strong evidence that the model has internalized a structured world model capable of state estimation.

\subsection{Metrics\label{subsection-metrics}}
Given an action sequence $s = (a_1, \dots, a_n)$ and its corresponding true state $S = \delta(S_0, s)$, we prompt the model to generate a predicted state $\tilde{S}$ using $s$ as an input. We evaluate the model's performance by comparing $S$ and $\tilde{S}$ using a suite of metrics designed to capture both syntactic accuracy and semantic fidelity.
\paragraph{State based metrics}

To better capture semantic correctness, we define metrics based on the sets of valid action sequences under a given state. Let $\mathcal{A}_S^m$ denote the set of all valid action sequences of length $m$ starting from state $S$.
We compare the sets $\mathcal{A}_S^m$ and $\mathcal{A}_{\tilde{S}}^m$ using the precision/recall formulation in Appendix \ref{appendix}.
In practice, computing these sets exactly is infeasible due to their exponential size in $m$. 
However, uniform sampling from $\mathcal{A}_S^m$ is itself intractable. Instead, we approximate this via \textbf{uniform branch sampling}: at each step $i$, we sample an action $a_i$ uniformly from the valid set $\Sigma_{S_{i-1}}$ and apply it via the transition function to obtain $S_i = \delta(S_{i-1}, a_i)$. Repeating this $m$ times yields a trajectory $s = (a_1, \dots, a_m)$.
Let $U_b(S)$ denote the distribution over such sequences. We then define approximate precision and recall as:
$$p_m(S,\tilde{S})=\mathbb{E}_{s \sim U_b(\tilde{S})} \left[ \mathbf{1}_{\mathcal{A}_S^m}(s) \right]$$
$$r_m(S,\tilde{S})=\mathbb{E}_{s \sim U_b(S)} \left[ \mathbf{1}_{\mathcal{A}_{\tilde{S}}^m}(s) \right]$$

These quantities reflect how well the predicted state $\tilde{S}$ preserves the behavior of the true state $S$, in terms of valid action trajectories. While state-based metrics are more faithful to the underlying semantics of state prediction, they are computationally expensive and depend on the trajectory length $m$. In practice, $m$ can be selected based on task complexity or evaluation constraints. Despite their cost, these metrics provide a much richer and more actionable signal than simpler string-based comparisons.
\vspace{-0.5cm}
\paragraph{Expected Values}

Consider a simplified case where the tree of possible action sequences generated from a state $S$ is \emph{homogeneous}, meaning that at each node, the proportion of child nodes corresponding to valid continuations under the state $\tilde{S}$ is constant. Let $p$ denote this proportion, i.e., the probability that a randomly selected legal action from any state results in a sequence accepted by $\tilde{S}$.

In this setting, since each step in the sequence independently maintains a success probability $p$, the expected probability that a full sequence of length $m$ is accepted by $\tilde{S}$ is 
$ p_m = p^m$. This analysis reveals that $p_m$ decays exponentially with the sequence length $m$ in the homogeneous case. Moreover, even in non-homogeneous trees, if the local acceptance probability at each step, denoted $p_s = \mathbb{P}(s_{i} \in \mathcal{A}_{\tilde{S}}^1 \mid s_{<i})$, is uniformly bounded above by some constant $M < 1$, then $p_m$ still decays exponentially as 
$p_m \leq M^m$. This exponential behavior highlights a fundamental challenge in estimating long-horizon compatibility between predicted and true states, motivating the need for careful design of sampling strategies and smoothing techniques in practice.

\section{Sampling algorithm}

\paragraph{Sample Complexity Analysis}

In our estimation procedure, we sample $N$ sequences of length $m$ from a given state. To ensure that the standard error of our estimate is of the same order of magnitude as the mean $p_m$, we analyze the variance of the binary indicator variable $\mathbf{1}_{\mathcal{A}_{\tilde{S}}^m}(s)$, which follows a Bernoulli distribution with parameter $p_m$.

The variance is $p_m (1 - p_m)$ while the standard error (SE) of the mean over $N$ independent samples is 
$
\text{SE} = \sqrt{\frac{p_m (1 - p_m)}{N}} \approx \sqrt{\frac{p_m}{N}},
$
where the approximation assumes $p_m \ll 1$, which holds due to the exponential decay of $p_m$ with $m$.

To achieve a signal-to-noise ratio (mean divided by standard error) of order one, we require
$
\frac{p_m}{\text{SE}} \approx \sqrt{N p_m} \approx 1,\text{ implying that }N \approx \frac{1}{p_m}$.

In the homogeneous tree case, we previously showed that the probability of a valid sequence of length \( m \) is given by
$
p_m = e^{-\lambda m},
$
for some constant \( \lambda = -\log p > 0 \), where \( p \) is the branching probability. Substituting this into our sample complexity estimate, we obtain
$
N = O(e^{\lambda m}).
$
This result implies that the sample complexity of our estimation procedure grows exponentially with the sequence length \( m \).

\subsection{Intermediate probability estimator}

Because a naive sampler must run for an exponential increasing amount of shots with respect to the depth we look at using an estimator based on conditional probabilities $p(s'\in \mathcal{A}_{\tilde{S}}^m|s\in\mathcal{A}_{\tilde{S}}^m)$ where $s'$ is $s$ with a random action appended. Note that we call this quantity $p_s$. Now we look for a certain $m$ at $\mathbb{E}_{s \sim U_b(S)}(p_s)=v_m$ (important $p_m\neq v_m$). From this we reconstruct $p_m=\Pi_{i=1}^m v_i$. In the case of an homogeneous tree the expectation of this product is $p^m$ as $v_m=p$. Here the error for each $v_i$ is $(1-v_i)\sqrt{\frac{v_i}{N}}$. If we maintain $(1-v_i)\sqrt{\frac{v_i}{N}}<<v_i$ the total error can be approximated to $\sum_{i=1}^m(1-v_i)\sqrt{\frac{v_i}{N}}$. If we fall back to an homogeneous tree then the error is $m(1-p)\sqrt{\frac{p}{N}}$. From this we get $N=O(m^2(1-p)^2p)$. Here the number of necessary shots increases quadratically with $m$ instead of exponentially.

\section{Experimental Setup}

\subsection{Metrics validation}

\begin{figure*}
    \centering
    \begin{minipage}[t]{0.49\linewidth}
        \centering
        \includegraphics[width=\linewidth]{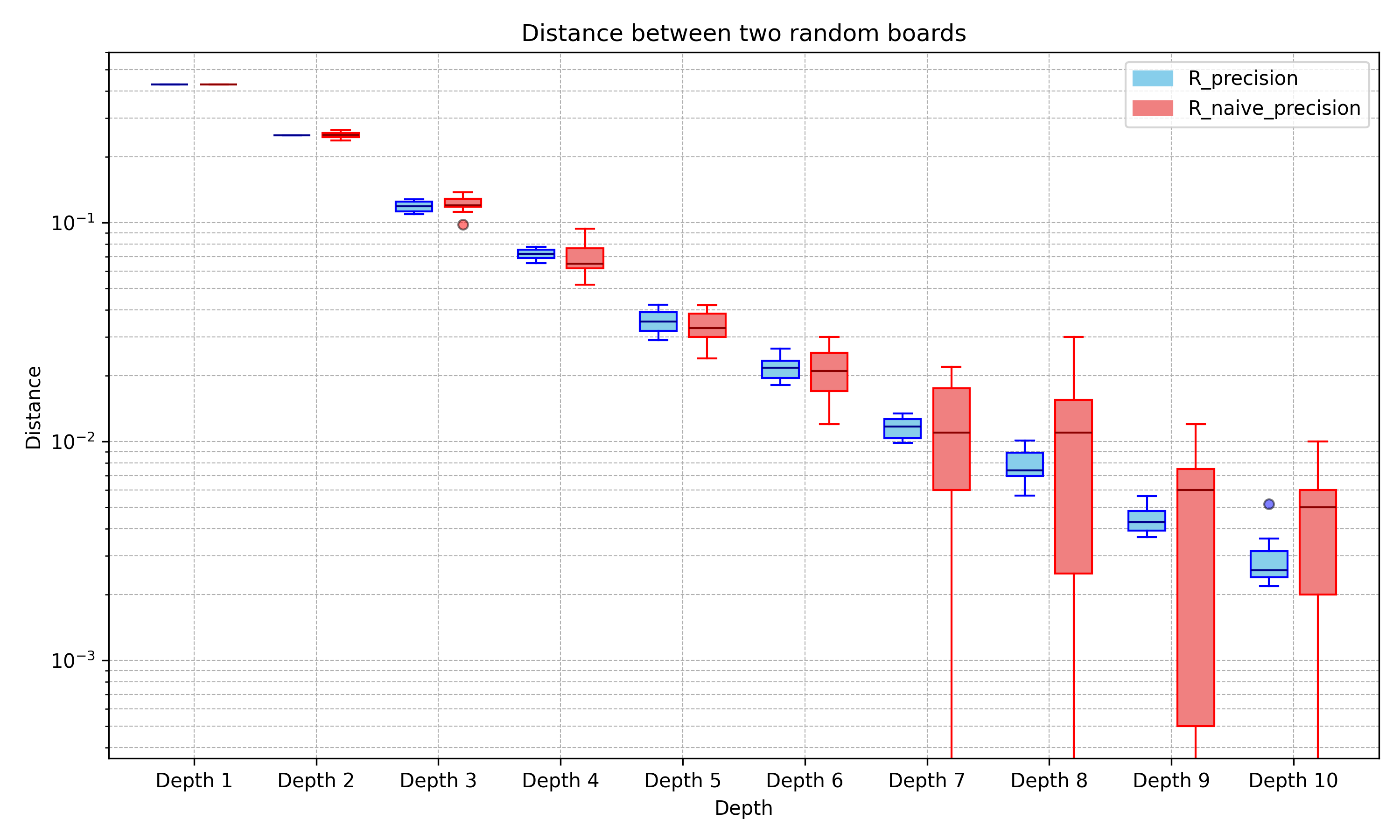}
        \caption{\(p_m\) as a function of \(m\) for two fixed states.}
        \label{fig:boxplotm}
    \end{minipage}
    \hfill
    \begin{minipage}[t]{0.49\linewidth}
        \centering
        \includegraphics[width=\linewidth]{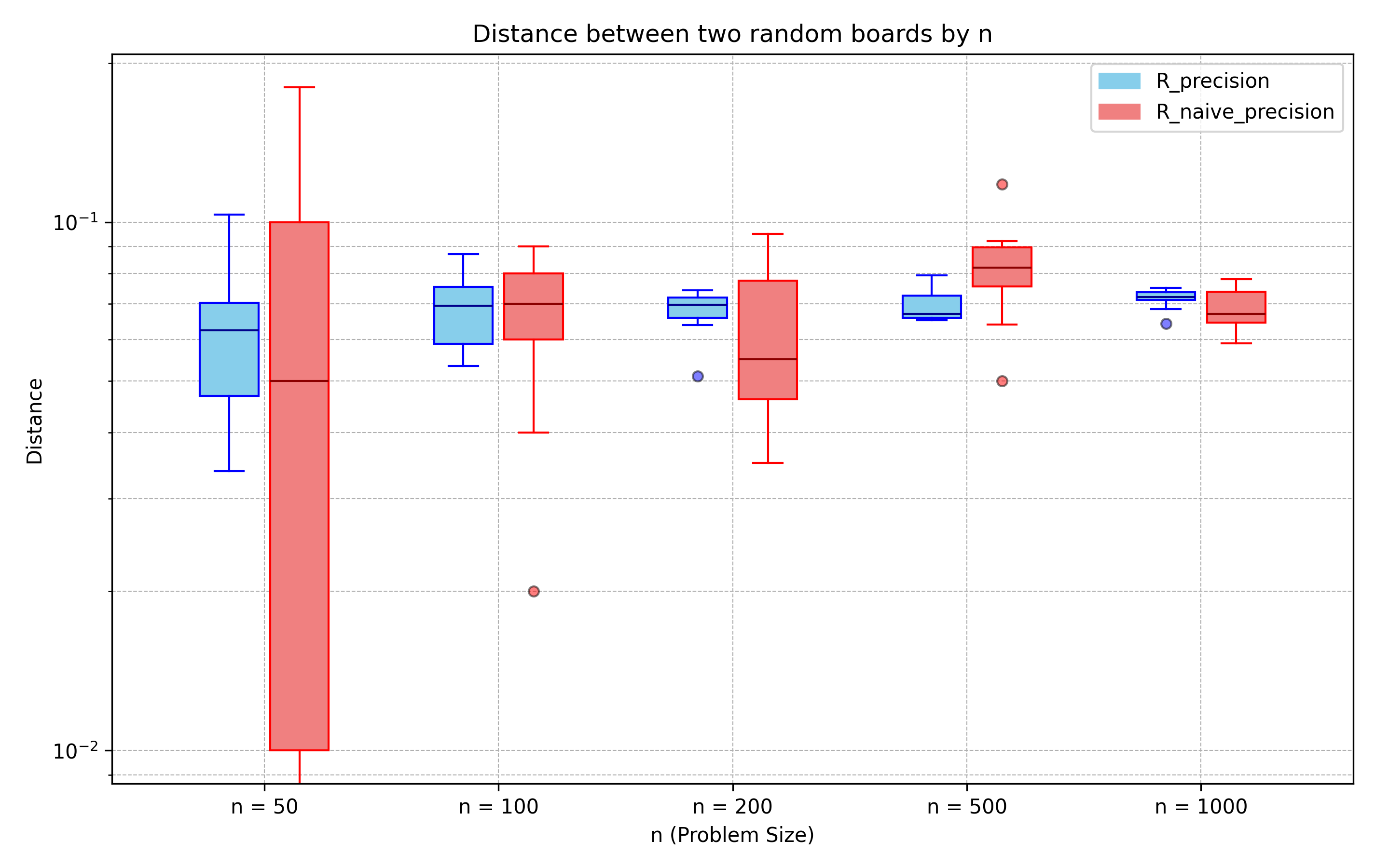}
        \caption{\(p_m\) as a function of \(N\) for two fixed states at \(m=4\).}
        \label{fig:boxplotN}
    \end{minipage}
\end{figure*}

We first validate our proposed metrics by examining their behavior on two fixed chessboard positions derived from random move sequences. We analyze the evolution of \(p_m\) with increasing \(m\) and assess the variance across multiple runs in figure \ref{fig:boxplotm}.

With a fixed sample size \(N=500\), the variance grows with \(m\) on a log scale. As expected, the naive estimator becomes increasingly noisy as \(m\) increases, while our second method remains significantly more stable.

Next, we examine the impact of sample size \(N\) at a fixed depth \(m=4\) for the same states in figure \ref{fig:boxplotN}.

The variance of our second estimator decreases substantially faster than the naive method. This advantage grows with deeper sequences or more divergent states, where \(v_m\) approaches zero. For reference, the states differ by an edit distance of 17 and have zero exact matches.

%\color{blue}
\subsection{Comparison to previous metrics}

We look at the correlations between our proposed metrics and the edit distance (levenshtein distance). To do so, we select a sample of 10000 real chess games selected from the website Lichess. We then separate in groups of 2000 samples, for each we select a specific game length (from 5 to 50 moves) and we cut the game to the specified length (we make sure beforehand that the total games are longer). Then we give to openai's GPT4o model the pgn interpretation of the game as well as instructions on what it has to do (convert to the "FEN" standard board representation of the game). Then we use our metrics (we take depth $m=4$) as well as levenshtien distance to evaluate our outputs.

\begin{figure*}
    \centering
    \begin{minipage}[t]{0.49\linewidth}
        \centering
        \includegraphics[width=\linewidth]{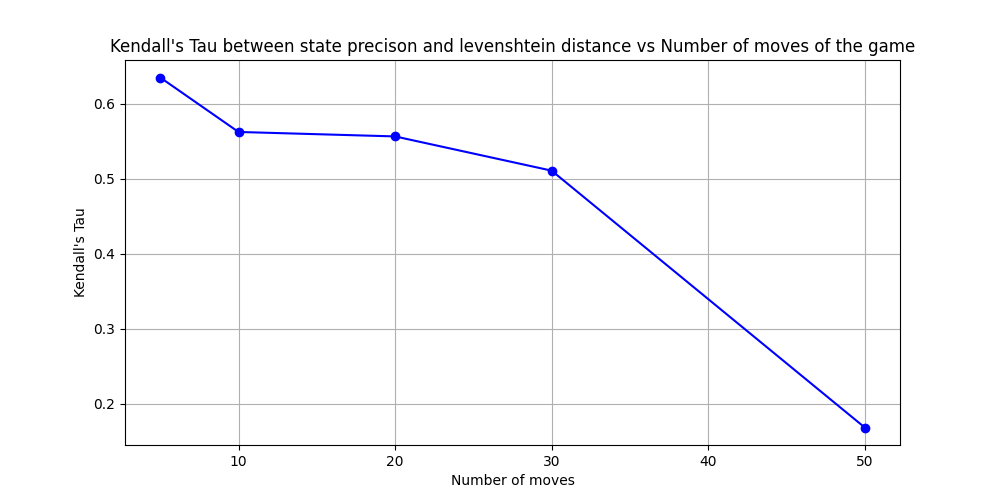}
        \caption{Values of Kendall's tau per group with varying number of moves per group}
        \label{fig:kendall}
    \end{minipage}
    \hfill
    \begin{minipage}[t]{0.49\linewidth}
        \centering
        \includegraphics[width=\linewidth]{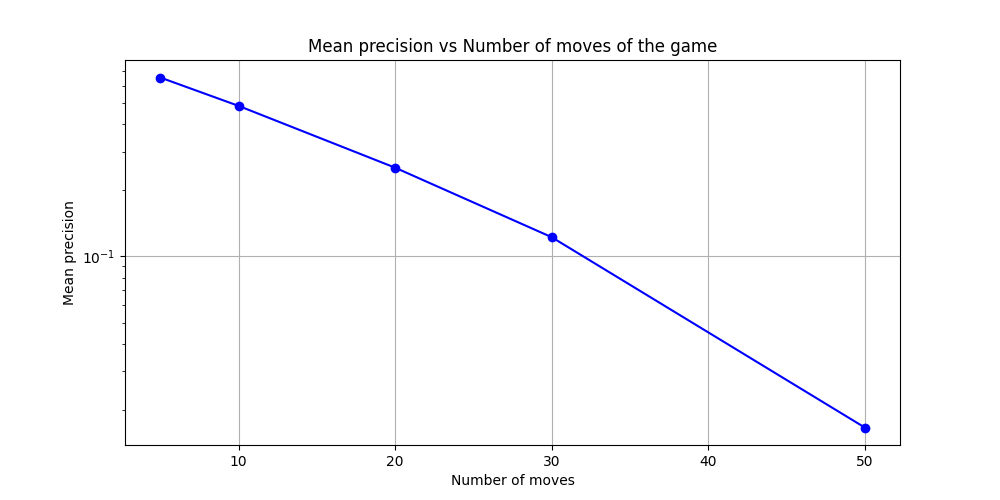}
        \caption{$p_4$ on average for each group (log scale, values from 0.6 to 0.015)}
        \label{fig:mean_precision}
    \end{minipage}
\end{figure*}

In order to tell how correlated our metrics are to the edit distance we look at the kendall's tau of the two distributions; state precision and $-1\times$ edit distance, first for the entirety of our sample and then by group. The overall kendall's tau is of 0.69, which can be interpreted as a strong correlation. However upon looking at values by groups we discover a different story, see figure \ref{fig:kendall}

We see that the correlation actually decreases with the number of moves of the game. This can be interpreted as when the number of moves increases the scores overall gets lower (see figure \ref{fig:mean precision}) and when they do the metrics we use become almost uncorrelated.

\color{black}
% \subsection{LLM benchmarking}
% We evaluate our metrics on 5,000 real human-played online chess games, divided into 10 groups. For each group, we select the first \(k\) moves (ranging from 5 to 50) and prompt GPT-4o to generate the current game state in FEN notation. We then apply our precision metric \(p_m\) with \(m=4\):

% \begin{figure}[h!]
%     \centering
%     \includegraphics[width=0.75\linewidth]{figures/pgn_Board_precision_GPT-4o.png}
%     \caption{\(p_m\) as a function of \(k\) for two fixed states with \(m=4\).}
%     \label{fig:enter-label}
% \end{figure}

The sharp decline in \(p_m\) as \(k\) increases highlights GPT-4o’s growing difficulty in accurately reconstructing the board state. Notably, the probability of producing a legal next move remains near one, indicating that the model still captures some aspects of valid gameplay. For comparison, the probability of a legal next move on random boards after 5 moves is approximately \((8 \pm 2) \times 10^{-4}\), whereas GPT-4o achieves an average \(p_m\) around 0.6. This demonstrates that while the model’s states are significantly better than random, performance degrades on longer sequences, reflecting challenges in both long-range state tracking and precise board reconstruction.

\section{Discussion}
A major limitation of our method is its sensitivity to the prompting strategy used to query the language model. Variations in phrasing, formatting, or the inclusion of intermediate reasoning steps (e.g., chain-of-thought prompting) can substantially affect the generated state representations. While this reflects a realistic deployment scenario, it complicates the interpretation of benchmarking results. Systematic studies on prompt robustness, prompt tuning, or self-verification techniques could help mitigate this. Another limitation is the assumption of access to a reliable and executable action model for the environment (e.g., the rules of chess), which may not be available or tractable in more open-ended or less formalized domains. Our approach also assumes that the ground-truth states are accurately labeled and that the representation space (e.g., FEN for chess) is sufficiently expressive to capture task-relevant differences. This assumption may not hold in domains with latent or ambiguous state representations. Additionally, although we framed the $m$-step precision/recall computation as a hyperparameter-dependent process, it introduces a trade-off between metric sensitivity and computational cost. Future work could explore strategies to average over multiple values of $m$, or replace it with an adaptive stopping criterion. We also note the possibility of using importance sampling or guided rollouts based on fitness functions or model confidence to further improve sampling efficiency. Finally, while we focused on the forward action space as the basis for our metric, a more comprehensive evaluation could also incorporate backward reasoning (e.g., whether the state is plausible given earlier context), or model uncertainty.

\section{Conclusion}
We propose a novel framework to evaluate language models' state-tracking abilities through metrics grounded in downstream task validity, rather than superficial representation similarity. Our key contributions include new evaluation metrics, notably \(p_m\) and \(r_m\), and efficient sampling-based estimators capable of handling exponentially large search spaces. Experiments on the chess domain demonstrate that our metrics provide sensitive and reliable assessments, revealing notable limitations in state reconstruction even for powerful models like GPT-4o as sequence lengths increase. Compared to traditional metrics such as exact match or edit distance, our approach better captures the semantic correctness of predicted states by considering their impact on valid subsequent actions. Despite its strengths, our method is sensitive to prompt design and computationally intensive for large depths \(m\). Future work includes addressing these challenges and extending the framework to other structured domains such as program synthesis, dialog tracking, and robotic planning.
Overall, this work offers a more principled, task-aware evaluation paradigm that aligns model assessment with downstream utility in structured reasoning tasks.

\bibliography{main}
\bibliographystyle{icml2025}

%%%%%%%%%%%%%%%%%%%%%%%%%%%%%%%%%%%%%%%%%%%%%%%%%%%%%%%%%%%%%%%%%%%%%%%%%%%%%%%
%%%%%%%%%%%%%%%%%%%%%%%%%%%%%%%%%%%%%%%%%%%%%%%%%%%%%%%%%%%%%%%%%%%%%%%%%%%%%%%
% APPENDIX
%%%%%%%%%%%%%%%%%%%%%%%%%%%%%%%%%%%%%%%%%%%%%%%%%%%%%%%%%%%%%%%%%%%%%%%%%%%%%%%
%%%%%%%%%%%%%%%%%%%%%%%%%%%%%%%%%%%%%%%%%%%%%%%%%%%%%%%%%%%%%%%%%%%%%%%%%%%%%%%
\newpage
\appendix
\onecolumn
\section{Appendix \label{appendix}}

\paragraph{Finite State Automaton}
\paragraph{State based metrics}

To better capture semantic correctness, we define metrics based on the sets of valid action sequences under a given state. Let $\mathcal{A}_S^m$ denote the set of all valid action sequences of length $m$ starting from state $S$.
We compare the sets $\mathcal{A}_S^m$ and $\mathcal{A}_{\tilde{S}}^m$ using a precision/recall formulation:
\[\text{Precision} = \frac{|\mathcal{A}_S^m \cap \mathcal{A}_{\tilde{S}}^m|}{|\mathcal{A}_{\tilde{S}}^m|}, \quad
\text{Recall} = \frac{|\mathcal{A}_S^m \cap \mathcal{A}_{\tilde{S}}^m|}{|\mathcal{A}_S^m|}
\]
In practice, computing these sets exactly is infeasible due to their exponential size in $m$. To address this, we express these quantities as expectations over indicator functions:
\[
\text{Precision} = \mathbb{E}_{s \sim \mathcal{U}(\mathcal{A}_{\tilde{S}}^m)}\left[\mathbf{1}_{\mathcal{A}_S^m}(s)\right], 
\]\[
\text{Recall} = \mathbb{E}_{s \sim \mathcal{U}(\mathcal{A}_S^m)}\left[\mathbf{1}_{\mathcal{A}_{\tilde{S}}^m}(s)\right]
\]
However, uniform sampling from $\mathcal{A}_S^m$ is itself intractable. Instead, we approximate this via \textbf{uniform branch sampling}: at each step $i$, we sample an action $a_i$ uniformly from the valid set $\Sigma_{S_{i-1}}$ and apply it via the transition function to obtain $S_i = \delta(S_{i-1}, a_i)$. Repeating this $m$ times yields a trajectory $s = (a_1, \dots, a_m)$.
Let $U_b(S)$ denote the distribution over such sequences. We then define approximate precision and recall as:
$$p_m(S,\tilde{S})=\mathbb{E}_{s \sim U_b(\tilde{S})} \left[ \mathbf{1}_{\mathcal{A}_S^m}(s) \right]$$
$$r_m(S,\tilde{S})=\mathbb{E}_{s \sim U_b(S)} \left[ \mathbf{1}_{\mathcal{A}_{\tilde{S}}^m}(s) \right]$$

These quantities reflect how well the predicted state $\tilde{S}$ preserves the behavior of the true state $S$, in terms of valid action trajectories. While state-based metrics are more faithful to the underlying semantics of state prediction, they are computationally expensive and depend on the trajectory length $m$. In practice, $m$ can be selected based on task complexity or evaluation constraints. Despite their cost, these metrics provide a much richer and more actionable signal than simpler string-based comparisons.

\begin{itemize}
    \item $\mathcal{S}$ is the finite set of valid states.
    \item $\mathcal{S}^* = \mathcal{S} \cup \{0\}$ is the augmented state space, including a special error (or sink) state $0$.
    \item $\Sigma$ is the finite input alphabet (i.e., the set of actions).
    \item $S_0 \in \mathcal{S}$ is the initial state.
    \item $\delta : \mathcal{S}^* \times \Sigma \rightarrow \mathcal{S}^*$ is the transition function.
\end{itemize}

The transition function is defined as:
\[
\delta(S, a) = 
\begin{cases}
    S' \in \mathcal{S} & \text{if } a \in \Sigma_S \text{ and } S \neq 0, \\
    0 & \text{if } a \notin \Sigma_S \text{ or } S = 0.
\end{cases}
\]

This construction ensures that $\delta$ is total, i.e., it produces a well-defined output for all pairs $(S, a) \in \mathcal{S}^* \times \Sigma$. Once the automaton transitions to the error state $0$, it remains there for all subsequent actions:
\[
\delta(0, a) = 0 \quad \forall a \in \Sigma.
\]

\paragraph{Exact Match}
We measure the probability that the predicted state exactly matches the ground-truth state:
\[
\mathbf{p}(\tilde{S} = S)
\]
This metric evaluates whether the model can perfectly reconstruct the correct state from the input sequence. It is strict and binary, any deviation from the target is counted as an error. As a result, low scores under this metric do not convey how close the predicted state is to the correct one, limiting its informativeness.

\paragraph{Edit distance}

To provide a more graded notion of correctness, we use the Levenshtein distance $\text{lev}(S, \tilde{S})$ between the textual representations of the true and predicted states. This measures the minimum number of single-character edits (insertions, deletions, or substitutions) required to transform one string into the other.
Since $\text{lev}(S, \tilde{S}) \in \mathbb{N}$ and can be unbounded, we normalize it into the $[0,1]$ range using an exponential kernel:
\[
\mathbb{E}\left[e^{-\lambda \cdot \text{lev}(S, \tilde{S})}\right]
\]
where $\lambda > 0$ is a hyperparameter. When $\lambda$ is large, the metric behaves similarly to exact match; when small, it becomes insensitive to differences. A drawback of edit distance is that it treats all changes equally, regardless of their semantic impact; i.e., how a change affects the resulting valid action set $\Sigma_S$ is not considered.
%%%%%%%%%%%%%%%%%%%%%%%%%%%%%%%%%%%%%%%%%%%%%%%%%%%%%%%%%%%%%%%%%%%%%%%%%%%%%%%
%%%%%%%%%%%%%%%%%%%%%%%%%%%%%%%%%%%%%%%%%%%%%%%%%%%%%%%%%%%%%%%%%%%%%%%%%%%%%%%

\begin{algorithm}[t]
\caption{Intermediate Probability Estimation}
\begin{algorithmic}[1]
\Require $N$: maximum list size, $m$: trajectory depth, $s_1$: starting state, $s_2$: comparison FSA
\Ensure Approximate $\mathbb{P}_{s \sim U_b(s_1)}[s \in \mathcal{A}_{s_2}^m]$ by computing $\prod_{i=1}^kv_k$ iteratively
\State $L \gets [(s_1, 1)]$ 
\For{$i = 1$ to $m$}
    \State $L' \gets [\ ]$ 
    \ForAll{$(j, w) \in L$}
        \State $\text{new} \gets \texttt{Legal\_moves\_add}(j)$ \Comment{Get legal next states}
        \State $w' \gets w / |\text{new}|$ 
        \ForAll{$m \in \text{new}$}
            \If{$s_2.\texttt{accepts}(m)$}
                \State $L'.\texttt{append}((m, w'))$
            \EndIf
        \EndFor
    \EndFor
    \If{$|L'| > N$}
        \State $L' \gets \texttt{sample}(L', N)$ \Comment{During sampling weights are rescaled}
    \EndIf
    \State $L \gets L'$
\EndFor
\State \Return $\sum_{(m, w) \in L} w$
\label{algo2}
\end{algorithmic}
\end{algorithm}

\subsection{Naive algorithm}

\begin{algorithm}[H]
\caption{Naive Precision/Recall Estimation}
\begin{algorithmic}[1]
\Require $N$: maximum number of sequences, $m$: depth, $s_1$: initial state, $s_2$: comparison FSA
\Ensure Approximate $\mathbb{P}_{s \sim U_b(s_1)}[s \in \mathcal{A}_{s_2}^m]$
\State $L \gets [s_1]$ %\Comment{List of current sequences}
\For{$i = 1$ to $m$}
    \State $L' \gets [\ ]$ %\Comment{New list of extended sequences}
    \ForAll{$j \in L$}
        \State $L' \gets L' \cup \texttt{Legal\_moves\_add}(j)$ %\Comment{Extend each sequence}
    \EndFor
    \If{$|L'| > N$}
        \State $L' \gets \texttt{sample}(L', N)$ %\Comment{Sample to control list size}
    \EndIf
    \State $L \gets L'$
\EndFor
\State $K \gets |L|$
\State $A \gets 0$ %\Comment{Accepted sequence counter}
\ForAll{$\text{seq} \in L$}
    \If{$s_2.\texttt{accepts}(\text{seq})$}
        \State $A \gets A + 1$
    \EndIf
\EndFor
\State \Return $A / K$
\end{algorithmic}
\end{algorithm}

\end{document}